\documentclass{article}

\usepackage[T1]{fontenc}
\usepackage{amssymb,amsmath,amsthm}
\usepackage{txfonts}
\usepackage{microtype}
\usepackage{relsize}
\usepackage{todonotes}
\usepackage{amsfonts}
\usepackage{mathtools}
\usepackage{commath}
\usepackage[sc,osf]{mathpazo}
\usepackage{mathtools}

\usepackage{authblk}

\usepackage{graphicx}
\usepackage{subfigure}

\usepackage{natbib}

\usepackage{algorithm}
\usepackage{algorithmic}

\usepackage{hyperref}
\usepackage{url}
\DeclareMathOperator{\EX}{\mathbb{E}} % expected value

\usepackage{mlp2017}
\mlptitlerunning{\studentNumber}
\title{Conditional GANs For Painting Generation}
\author{
    \textbf{Adeel Mufti, Biagio Antonelli, Julius Monello}\\
    The University of Edinburgh\\
}
\date{}

\begin{document}

\maketitle

\begin{abstract}
We examined the use of modern Generative Adversarial Nets to generate novel images of oil paintings using the Painter By Numbers dataset. We implemented Spectral Normalization GAN (SN-GAN) and Spectral Normalization GAN with Gradient Penalty, and compared their outputs to a Deep Convolutional GAN. Visually, and quantitatively according to the Sliced Wasserstein Distance metric, we determined that the SN-GAN produced paintings that were most comparable to our training dataset. We then performed a series of experiments to add supervised conditioning to SN-GAN, the culmination of which is what we believe to be a novel architecture that can generate face paintings with user-specified characteristics.
\end{abstract}
%Can you be more specific about your results and conclusions?

%Julius
\section{Introduction}
\label{sec:intro}

A motivation for this work can be given by Richard Feynman, who once stated: ``What I cannot create, I do not understand.'' Not only are Generative Adversarial Networks capable of creating entirely new data (such as images, text, or speech), they do so in an adversarial manner. This adversarial training enables the generative portion of the network to continually improve its representation of the true data distribution (that it attempts to capture and generate) until its outputs are indistinguishable from the real data. The uses for generative models are endless, ranging from image denoising, to video generation, to speech synthesis. Their generative process can be thought of similar to the way the human brain dreams, imagines, and predicts. Because adversarially trained generative models parameterize real world data in order to generate new samples, they tend to learn efficient, alternate representations of the data, which can be said to have a deeper understanding of the data.

The ultimate objective of our work was to train a GAN conditioned on different attributes of paintings in order to generate novel paintings that have specified attributes of our choosing. We believe that successful implementations of conditional GANs could have a number of interesting applications in new and not heavily explored areas. For example, they could be used by artists and other creatives to aid in the idea generation process for new projects, perhaps based on their own previous work and work of other artists. %(e.g. by conditioning on landscapes, GANs will generate a number of new landscapes, some of them could pick styles for several different artists and provide inspiration for new works).  

For our work, we first sought to improve upon baseline results previously achieved: since our initial experiments run using the Deep Convolutional GAN (DC-GAN) \cite{radford2015unsupervised} architecture did not produce `paintings' that were close to human indistinguishable from the training data, we set out to explore recent state-of-the-art GAN architectures in order to improve our results. This was especially important due to the fact that conditioning our GAN on image attributes effectively reduces the number of training examples, as the data are now split per attribute. 

More specifically, our initial goal was to create a GAN using the techniques and architecture detailed by \citet{miyato2018spectral} and measure its improvement on our baseline. From there, in a further attempt to improve our generated images, we added an additional ``gradient penalty" as specified by \citet{gulrajani2017improved}. We then aimed to add conditioning to our best performing GAN architecture, as measured quantitatively by the sliced Wasserstein distance (SWD) between generated and real images, as well as qualitatively via our own perception. 

We focused on paintings in 4 categories while improving our baselines: landscapes, flowers, portraits, and pre-processed portraits (which we will call faces) -- again taken from the Painter by Numbers (PBN) dataset \cite{duck2016painter}. We chose to condition the paintings -- specifically the faces -- on characteristics specified by Microsoft's Facial Recognition API\footnote{https://azure.microsoft.com/en-gb/services/cognitive-services/face/}. In order to improve the final quality of our generated conditional images, we additionally used portraits taken from the Behance Artistic Media (BAM) dataset \cite{Wilber_2017_ICCV}.

The structure of this paper is as follows. We first detail the theory behind recent related works in Sec.~\ref{sec:relatedwork}. From there, we describe the methodological underpinnings of our experiments in Sec.~\ref{sec:method}, after which we discuss in detail our experiments and results in Sec~.\ref{sec:exp}. In Sec.~\ref{sec:conclusions}, we summarise our experimental results and remark on our findings.

\section{Related Work}
\label{sec:relatedwork}

Generative Adversarial Nets were proposed by \citet{goodfellow2014generative} and consist of 2 neural networks with adversarial goals: a discriminative model D and a generative model G. D is trained on images from a `real' distribution, and G, given random (usually uniformly drawn) noise $\mathbf{z}$, generates new images in an attempt to trick D into classifying the images as emanating from the real distribution. Thus, discriminator D's goal is to give probabilities near $0$ to images generated by G, which we will call $G(z)$; generator G's goal is to get D to assign its images probabilities near $1$. This amounts to a minimax game, with the goal of the entire network to optimize the following loss function:
\begin{equation}
\begin{split}
    \min_{G}\max_{D} L(D,G) &= \mathbb{E}_{x\sim p_r}\left[\log(D(x)\right] \\
    &+ \ \mathbb{E}_{x\sim p_g}\left[\log(1-D(x))\right],
\end{split}
\end{equation}

where $p_r$ is the real data distribution, $p_g$ is the generator's learned distribution, and $D(x) \in [0,1]$ is the classification by the discriminator. At the optimal value for D, this loss function can be rewritten to be in terms of the distance measure known as Jensen-Shannon (JS) divergence between the real and generated distributions \cite{goodfellow2014generative}. 

This original formulation of GANs (using the JS divergence) was certainly groundbreaking, but suffers from many problems during training, most notable of which are:
\begin{itemize}
    \item Mode collapse: the generator G learns a setting of parameters wherein it only produces one image. This successfully fools the discriminator, but prevents G from learning the extent of the real data distribution.
    \item Vanishing gradient: as the discriminator gets better at discerning real versus generated images, then $D(x) \rightarrow 1$ for images $x$ drawn from $p_r$ and $D(x) \rightarrow 0$ for $x \sim p_g$. This causes the gradient of Eq. 1 to go towards $0$, and in the optimal case where D is never wrong, the gradient is $0$. So, if D improves too much, then learning slows significantly or even stops. And on the other hand, if D does not improve much, then G cannot learn to make better images. 
    \item No clear evaluation metric: since the true data distribution $p_r$ is not known, and also calculating the full distribution of $p_g$ is often not possible (as with other generative models), there is no one quantitative metric by which to evaluate the generated images \cite{theis2015note}.
\end{itemize}
%TODO: maybe add the bullet about low dimensional supports, since this the main thing WGAN fixes
\citet{salimans2016improved} suggested a series of improvements to account for these issues, notably mini-batch discrimination (wherein D looks at batches of generated images rather than one at a time) to help with mode collapse, and label smoothing, wherein the predictions of D are restricted so as not to get too close to 1, in order to help the vanishing gradient problem. \citet{arjovsky2017towards} suggested adding Gaussian noise to the inputs of D to further help with the vanishing gradient problem. In a followup paper, \citet{arjovsky2017wasserstein} proposed using the Wasserstein (also known as Earth-Mover) distance as a metric by which to evaluate the similarity between $p_r$ and $p_g$:
\begin{equation}
    W(p_r,p_g)=\sup_{\norm{f}_L\leq 1} \mathbb{E}_{x\sim p_r}\left[f(x)\right] - \mathbb{E}_{x\sim p_g}\left[f(x)\right]
\end{equation}

where $f$ is the discriminator (which maps from the real image space to $\mathbb{R}$) as long as $\norm{f}_L \leq 1$  is satisfied. $\norm{f}_L \leq 1$ enforces $f$ to be 1-Lipschitz continuous. While the full consequences of 1-Lipschitz continuity are outside the scope of this report, the main result is that the Wasserstein distance between $p_r$ and $p_g$ is constrained to be continuous and differentiable nearly everywhere, with a well-behaved gradient. Eq. 2 can be reformulated into a loss function, which forms the basis of the Wasserstein GAN \cite{arjovsky2017wasserstein}:
\begin{equation}
    L(p_r,p_g)=\max_{w\in \mathcal{W}} \mathbb{E}_{x\sim p_r}\left[f_w(x)\right] - \mathbb{E}_{z\sim p(z)}\left[f_w(G(z))\right].
\end{equation}
Here, the discriminator learns a function $w$ that is 1-Lipschitz continuous in order to compute the Wasserstein distance. The loss decreasing during training corresponds to a reduction in the Wasserstein distance between $p_r$ and $p_g$, and the generator produces more realistic images. 

In order to enforce the 1-Lipschitz constraint on the discriminator, \citet{arjovsky2017wasserstein} simply clipped the weights of the discriminator to small values, while admitting that their solution has major faults and could certainly be improved upon. Notably, \citet{gulrajani2017improved} point out that without taking special care to tune the bounds of the weight clipping, convergence can be very slow, and the discriminator could end up learning an overly simple function.

% Biagio up to conditioning, Adeel conditioning
\section{Methodology}
\label{sec:method}

To improve upon prior baselines using DC-GAN, before moving onto conditioning, we explored two recently proposed methods of enforcing the 1-Lipschitz constraint on the discriminator. 

\subsection{Spectral Normalization}
We first explored Spectral Normalization, which also enforces the 1-Lipschitz constraint by acting on the weights of the discriminator. However, rather than simply clipping the weights in each layer to be small, \citet{miyato2018spectral} normalize the spectral norm of the weight matrix $W$ at each layer in the discriminator. Let $\sigma(W)$ denote the largest singular value of $W$. \citet{miyato2018spectral} then transform $W$:
\begin{equation}
    \bar{W}_{SN} \coloneqq \frac{W}{\sigma(W)},
\end{equation}
such that $\sigma(\bar{W}_{SN}) = 1$, thus satisfying the 1-Lipschitz constraint.

%It is known that for a fixed generator, the expression of the discriminator is: 
%\begin{equation}
%     D_G^*(x) =  \frac{q_{data}(x)}{q_{data}(x) + p_G(x)} = sigmoid(f^*(x))  
%\label{3}     
%\end{equation}

%where $f^*(x) = \log q_{\mathrm{data}}(x) - log p_G(x)$. Its gradient:
%\begin{equation}
%    \nabla_x f^*(x) = \frac{1}{q_{data}(x)} \nabla_x q_{data} - \frac{1}{p_G} \nabla_x p_G(x) 
%\end{equation}
%can be unbounded or not even defined. Therefore, some kind of %regularity condition is needed . 

We used a TensorFlow implementation of Spectral Normalization which was publicly available as Python code on GitHub\footnote{https://github.com/minhnhat93/tf-SNDCGAN}. This served as a starting point for quickly being able to get our experiments off the ground.

\begin{figure}[!h]
\vskip 5mm
\begin{center}
\subfigure [Generator.] {
	\includegraphics[width=0.2\textwidth]{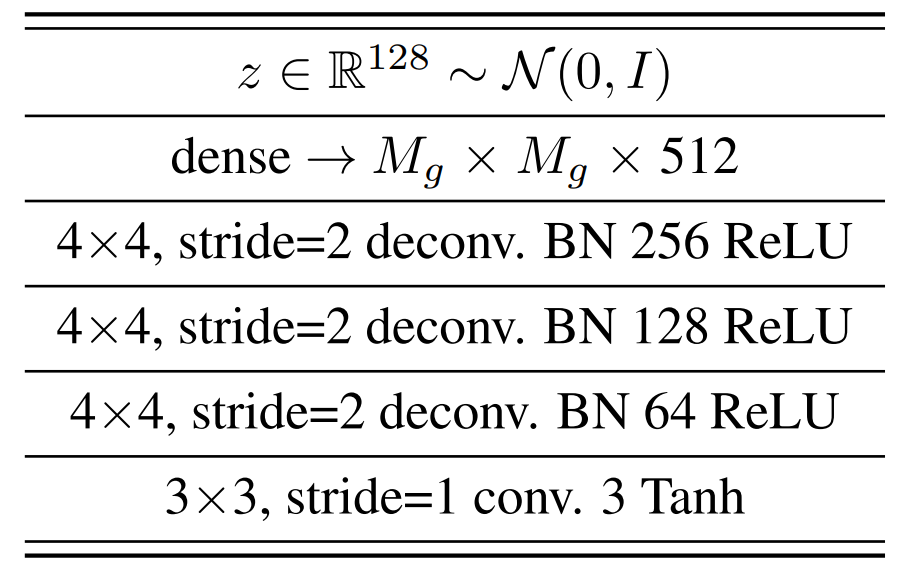}
}
\subfigure[Discriminator.]{
	\includegraphics[width=0.2\textwidth]{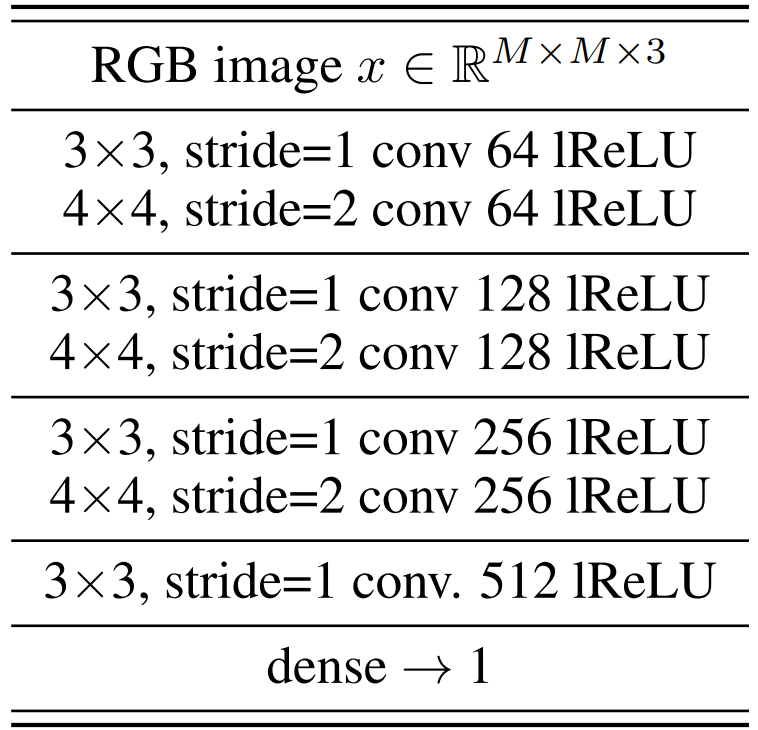}
}
\caption{SN-GAN architecture \citep{miyato2018spectral}.}
\label{fig:SNarch} 
\end{center}
\vskip -5mm
\end{figure}

The model architecture for Spectral Normalization GANs (SN-GANs) is illustrated in Fig.~\ref{fig:SNarch}. The slope of all lReLU activations is set to 0.1.

In an attempt to further improve our result, as suggested by  \citet{miyato2018spectral}, we then sought to add a secondary method to enforce 1-Lipschitz continuity on the discriminator.

\subsection{Gradient Penalty}
\label{ssec:method_GP}
Introduced by \citeauthor{gulrajani2017improved} (\citeyear{gulrajani2017improved}), the gradient penalty enforces 1-Lipschitz continuity by adding a regularizing term to the cost function of the GAN, rather than altering the weights. Thus, to Eq. 3 a regularizing term that penalizes large gradients is added:

\begin{equation}
    \lambda \EX_{\hat{x} \sim p_{\hat{x}}} [(  \left\Vert\left(\nabla_{\hat{x}} D(\hat{x})\right) \right\Vert_2 - 1 )^2]    
\end{equation}

where $\lambda > 0 $ is a regularization coefficient and $\hat{x}$ is a weighted average between pairs of images in $p_r$ and $p_g$. It should be noted that the calculation of $\nabla_{\hat{x}} D(\hat{x})$ requires an additional forward and backward pass through the network, which we expect to increase training time.
%\begin{equation}
%    \hat{x} := \epsilon x + (1 - \epsilon)\Tilde{x} 
%\end{equation}
%with $\epsilon \sim U[0,1], x \sim p_{data}, \tilde{x} = G(z), z \sim p_z$.

Since spectral normalization and gradient penalty act on different parts of the discriminator (on layer weights and the gradient of the loss function, respectively), \citet{miyato2018spectral} hypothesized and then found that the two approaches could work in complement.

We added the gradient penalty loss function to our spectral normalization codebase, and ran experiments using spectral normalization (SN) as well as the combination of spectral normalization and gradient penalty (SNGP).

%Adeel
\subsection{Conditioning}
\label{ssec:methodology-conditioning}
Once our original DC-GAN baselines were sufficiently improved, we set out to condition GANs to generate paintings with specific attributes that can be manually specified. As an example, generating a landscape or a portrait (as specified by the user) from the same GAN. Or, generating paintings of faces with specific attributes such as gender, emotion, or hair color. 

In order to condition GANs, we needed labels for our data. For landscapes versus portraits, we used the genre labels provided in the Painter By Numbers metadata. For portraits, we decided to use Microsoft Face API, which is a Machine Learning API that can take images of persons as input, and return a JSON of facial attributes such as age, hair color, gender, emotion, facial hair, and more. We extracted images of faces from the portraits genre in Painter By Numbers, and gathered attributes for each one using Face API. However, these did not prove to be sufficient during our conditioning experiments (see Sec.~\ref{ssec:experiments-conditioning}), as we only had 3,269 extracted faces available. So we set out to augment our faces dataset using the BAM dataset \citep{Wilber_2017_ICCV}, which provided us with 21,965 oil paintings of people (though not always with a clear face). Using OpenCV\footnote{https://opencv.org}, we extracted 4,326 faces from the BAM paintings, and also got facial attributes using Microsoft Face API. This gave us a total of 7,595 face painting images with facial attributes as labels. Finally, we augmented our faces by flipping each one along the x-axis (horizontally), thus doubling our face painting images to 15,190 in total -- a number comparable to the number of landscapes and un-preprocessed portraits originally available to us.\footnote{Since, at this point in our experiments,  quantitative evaluation via SWD was no longer necessary, we believed that the addition of new, previously unused data did not compromise any prior results.}

Conditioning approaches for GANs use a new label vector, which is commonly referred to as the $\mathbf{y}$ vector in literature. This label vector contains more information about the data, such as certain attributes, that the GAN is trained on. It is necessary to have a $\mathbf{y}$ with the same dimensionality for every datapoint being used in training. For semi-supervised conditioning, $\mathbf{y}$ can be uniform noise with dimension equal to number of classes being conditioned. Or, $\mathbf{y}$ can be one-hot encoding of the desired class labels for fully supervised conditioning \citep{DBLP:journals/corr/MirzaO14}. 

We decided to use one-hot encoding for supervised conditioning, so we could provide a manual one-hot encoded $\mathbf{y}$ vector (label) after training to produce a specific type of painting. For example, for landscapes and portraits, where $\mathbf{y}\in\mathbb{R}^{2}$, $\mathbf{y}$ was a one-hot encoding represented as [landscape, portrait] -- $\mathbf{y}=[1, 0]$ for all landscapes, and $\mathbf{y}=[0, 1]$ for all portraits. Similarly for conditioning on faces, we created $\mathbf{y}$ to be a one-hot encoding of the facial attributes gathered from Face API. For our experiments we first used all facial attributes provided through Face API (33 in total), one-hot encoded into a $\mathbf{y}$ vector. But after some difficulties in training (more details in Sec.~\ref{ssec:experiments-conditioning}), we limited these attributes to 6: gender (male when gender=1, female when gender=0), happiness, age from 0-9, black hair, blond hair, and facial hair. We picked these attributes because they offered a large number of samples available in our augmented faces data-subset, and because we thought they provided easily discernible visual distinctions.

There are a number of different conditioning approaches for GANs, meaning $\mathbf{y}$ can be introduced to both the discriminator and the generator during training in a variety of ways. `Vanilla' conditional GANs \citep{DBLP:journals/corr/MirzaO14} simply concatenate $\mathbf{y}$ to the noise vector $\mathbf{z}$ being input to the generator, and $\mathbf{y}$ to $\mathbf{x}$ being input to the discriminator, before $\mathbf{z}$ and $\mathbf{x}$ are input to the first layers of each respectively. For deep convolutional variants of GANs, some approaches concatenate $\mathbf{y}$ to $\mathbf{z}$ for the input to the generator in the same manner, but instead concatenate $\mathbf{y}$ to the dense layer at the end of all the convolutional layers \citep{gauthier2014conditional,reed2016generative}. Other variants, instead of putting $\mathbf{y}$ towards the end of the discriminator, add $\mathbf{y}$ at the beginning of the discriminator -- by either tiling $\mathbf{y}$ with each filter after the first convolutional layer \citep{DBLP:journals/corr/PerarnauWRA16}, or by adding $\mathbf{y}$ as an input to a dense layer that is reshaped to the height and width dimensions of the images and then concatenated as a fourth channel to each of the $\mathbf{x}$ (images input to the discriminator). 

We experimented on all types of conditional GANs mentioned above, and even combinations of the approaches, in an attempt to get conditioning to work. First we tried conditioning on MNIST \citep{lecun2010mnist} as a proof-of-concept for conditioning using the simple `vanilla' conditional GAN. For our painting image data-subsets, we experimented with conditioning using combined landscapes and portraits, baseline faces (from just Painter By Numbers), and eventually our augmented faces (Painter By Numbers + BAM) datasets. We believed that MNIST and then landscape + portraits provided the clearest class distinctions within the dataset, making it an easier conditioning task for the GAN. Once we got conditioning working for those, we moved on to experimenting with faces.

With each experiment, we observed the discriminator and generator loss numbers through the course of training. Additionally, our code produced an 8x8 grid of generated images as training progressed, which we also visually inspected. These allowed us to get an idea of how our experiment was progressing over the course of training. For the purpose of conditioning, we made our generated samples consist of an easily discernible 50-50 conditional split, depending on the data-subset we were training on. For example, for landscapes + portraits, we had the first half of the our 8x8 grid contain landscapes, and the second half portraits (the $\mathbf{y}$ passed into the generator for producing these samples had [landscape=1, portrait=0] for the first half and then [landscape=0, portrait=1] for the second half). For faces, we decided to split on the gender attribute, so if conditioning was working, we could observe the first half of our samples as faces of males, and second half as faces of females.

\subsection{Model Evaluation}
\label{ssec: model_eval}

In addition to visually comparing the results between our models as we sought to improve upon our DC-GAN baselines, we chose to use the Sliced Wasserstein Distance (SWD) as a quantitative evaluation measure. SWD is an efficient approximation to the Wasserstein distance, and effectively compares training and generated images in both appearance and variation at different resolutions \cite{karras2017progressive}. To do so, 100 images are generated and compared to 100 randomly selected images from the training data. It should be noted that often many more samples are taken in other literature, but this was not feasible due to limited computational resources. As such, our SWD numbers should not be compared to those from other papers, but are still appropriate to compare results of our own experiments. In our evaluations, we report one SWD per model per painting category, where the distances computed over resolutions of 128px, 64px, 32px, and 16px are averaged. A lower average SWD means the GAN performed better in generating images similar to the training data. 

It should also be noted that our training images were quite varied, even within classes. This is simply due to the nature of the PBN dataset, which consists of paintings (as opposed to many GAN papers which use heavily pre-processed, real photographs such as the CelebA dataset \cite{yang2015facial}). As such, sometimes when examining our model outputs and noting an example that, say, does not quite look like a landscape, the finding would be taken with a grain of salt because there were many similar paintings labeled as landscape in our training dataset.

%%%%%%%%%% EXPERIMENTATION PART %%%%%%%%%%%%%%%%%%%%%%%%%%%%%%%%%%%%%%%%%%%%%%%%%%%%%%%%%%%%%%%%%%%%%%%%%%%%%%%%%%%%%%%%%%%%%%%%%%%%%%%%%%%%%%%%%%%%%%%%%%%%%%%%

% 
\section{Experiments}
\label{sec:exp}

We trained both the SN and SNGP-GANs described above on subsets of the PBN dataset containing landscapes, flowers, portraits, and also faces extracted from portraits. The images for each category we experimented on were center-cropped and reduced to an equal height and width of 128px. The models were trained for 100,000 iterations -- we set the mini-batch size to 64, which means that our models were trained on 6.4 million images. All experiments used the Adam learning rule. SN-GAN experiments used a learning rate of 0.0002, $\beta_1 = 0.5$, and $\beta_2=0.999$ for Adam, while SNGP-GAN experiments had a learning rate of $5\cdot 10^{-5}$, $\beta_1 = 0$, and $\beta_2 = 0.9$ \cite{miyato2018spectral}. The regularization parameter for gradient penalty, $\lambda$, was set to be $1$.
%[Biagio: we could omit this]->Here we consider iteration and not anymore epochs, as it is a standard procedure in the GANs' literature and ensure that all the models are trained on the same number of images. 
\begin{figure}[!h]
\vskip 5mm
\begin{center}
\subfigure [DC-GAN] {
	\includegraphics[width=0.14\textwidth]{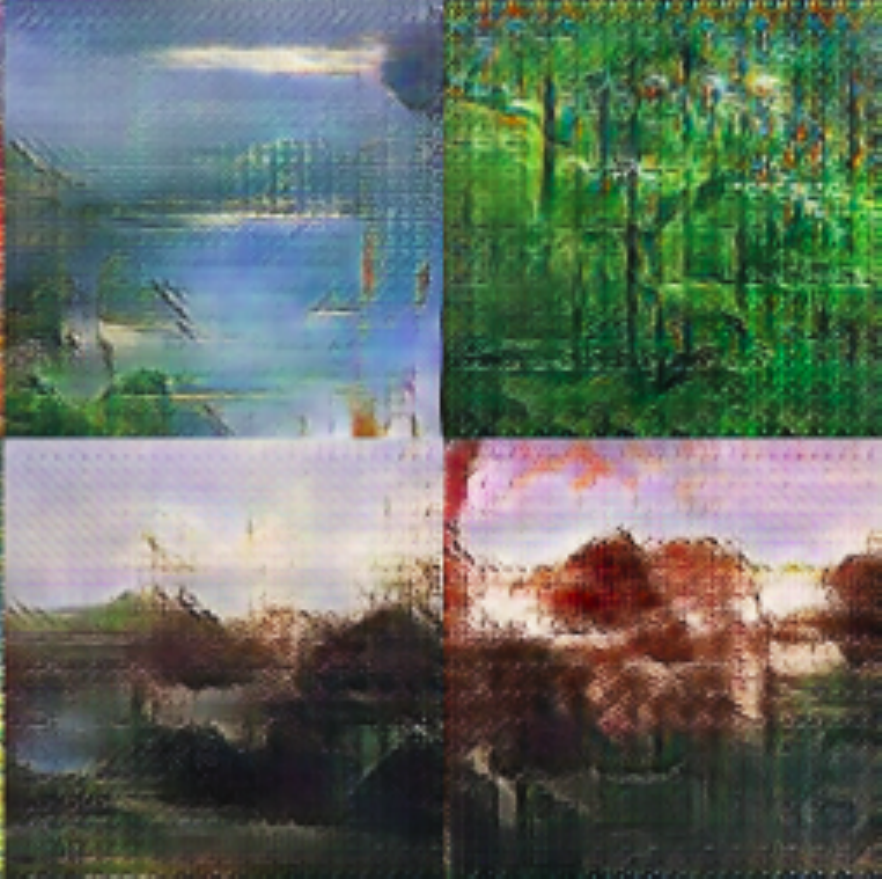}
}
\subfigure[SN-GAN]{
	\includegraphics[width=0.14\textwidth]{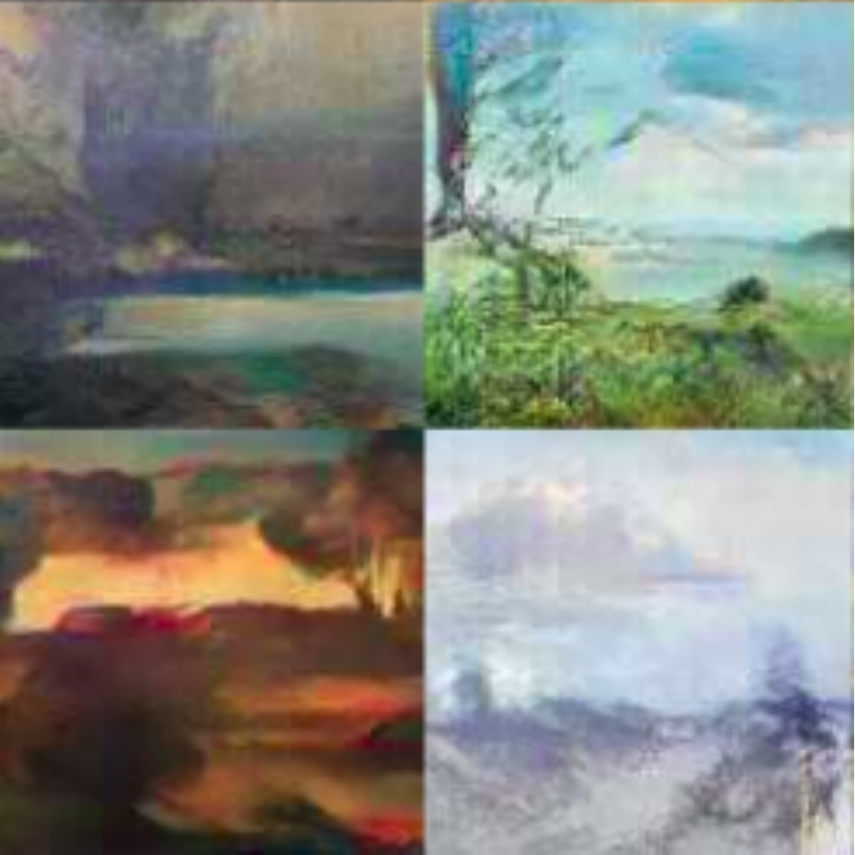}
}
\subfigure[SNGP-GAN]{
	\includegraphics[width=0.14\textwidth]{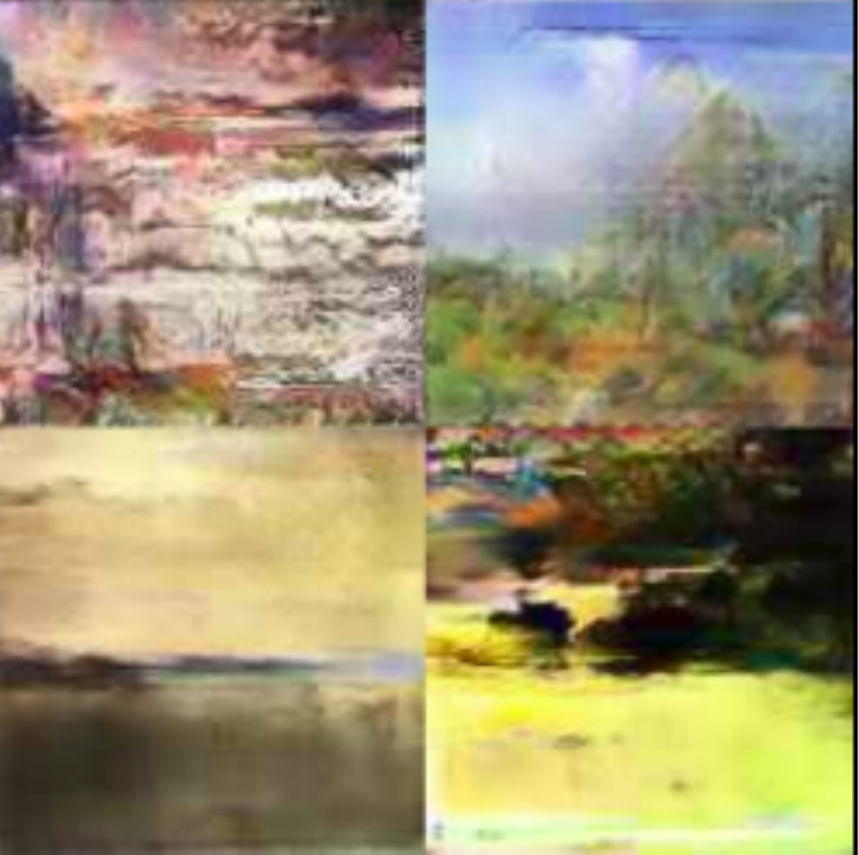}
}
\caption{Comparison of landscapes generated with DC-GAN, SN-GAN and SNGP-GAN.}
%. Spectral normalization significantly improves the quality of the generated images, while the addition of gradient penalty does not
\label{fig: comparison of landscapes}   
\end{center}
\vskip -5mm
\end{figure}

\subsection{Spectral Normalization}

Spectral normalization led to a significant increase in image quality on each of the image categories. Initially, results were self-evaluated visually. However, we compared the best model for each subcategory with SWD, as introduced in Sec. \ref{ssec: model_eval}.

\begin{figure}[!h]
\vskip 5mm
\begin{center}
\subfigure [DC-GAN] {
	\includegraphics[width=0.14\textwidth]{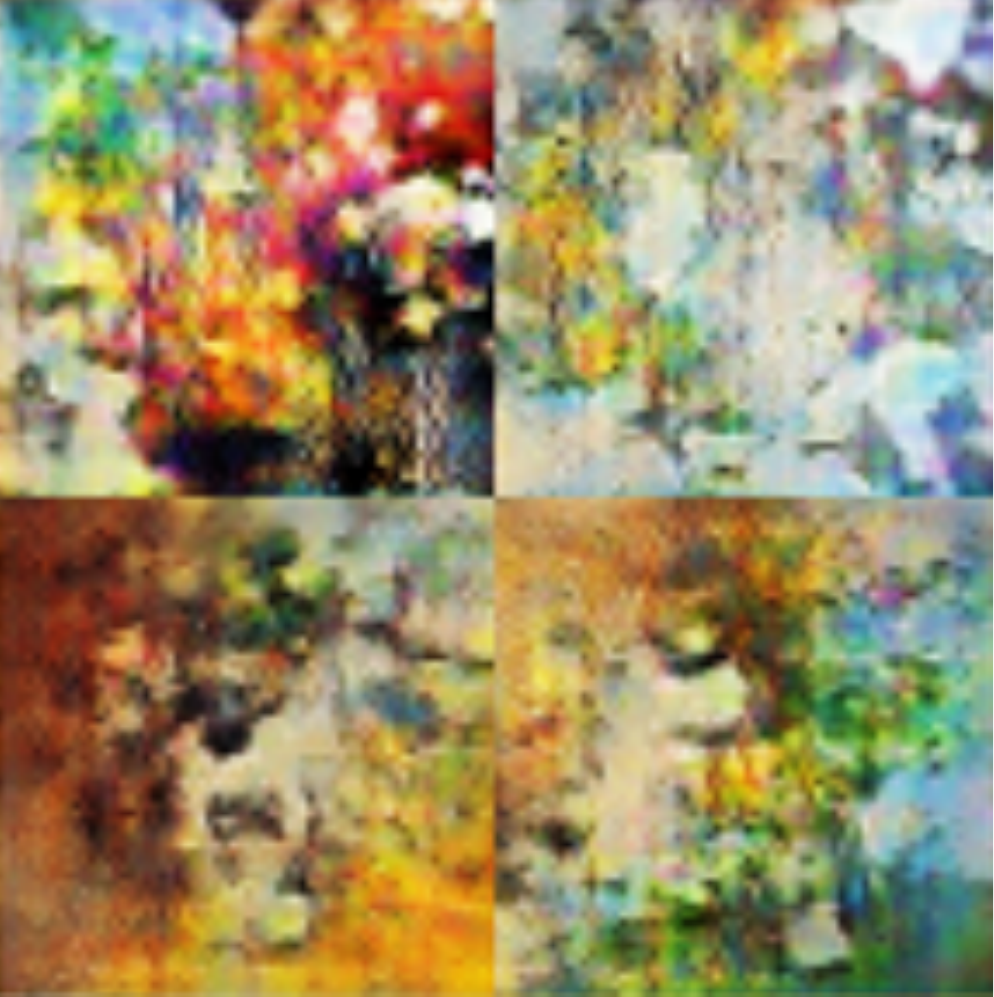}
}
\subfigure[SN-GAN]{
	\includegraphics[width=0.14\textwidth]{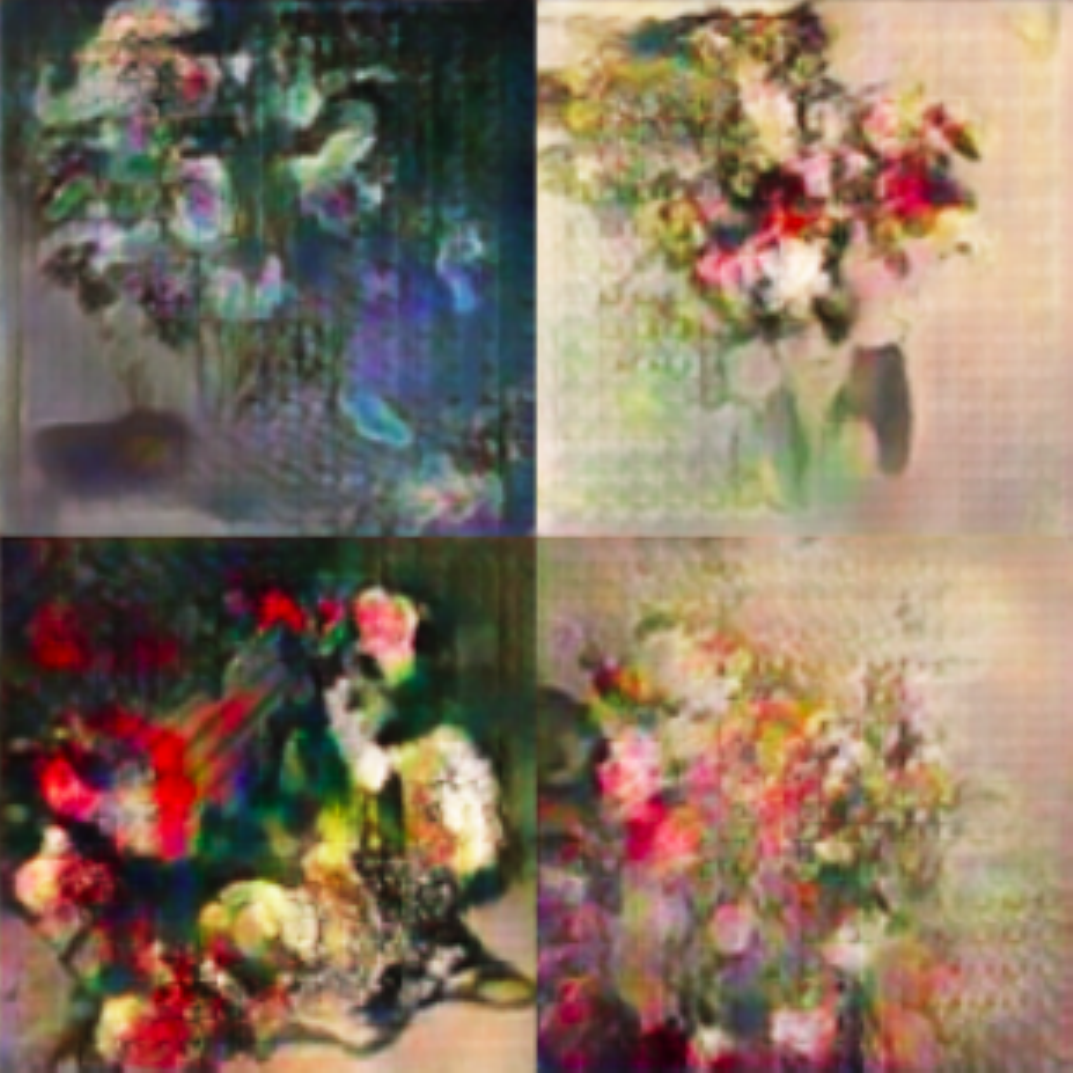}
}
\subfigure[SNGP-GAN]{
	\includegraphics[width=0.14\textwidth]{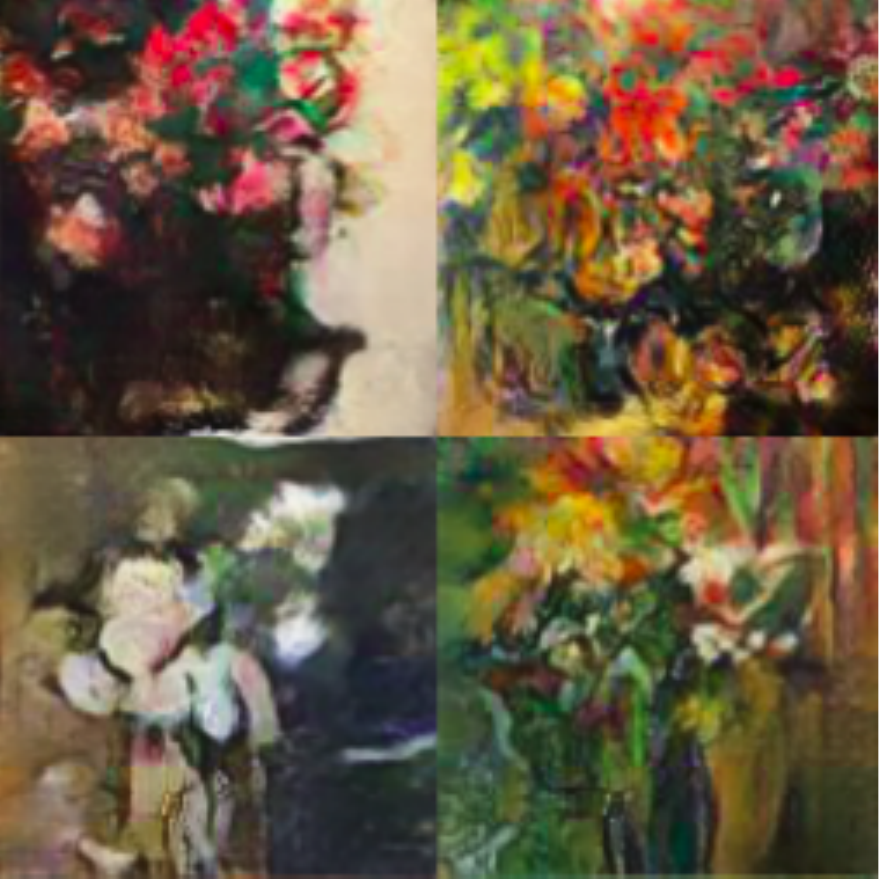}
}
\caption{Comparison of flowers generated with DC-GAN, SN-GAN and SNGP-GAN.}
\label{fig:comparison of flowers}   
\end{center}
\vskip -5mm
\end{figure}

For flowers, DC-GAN experiments performed were not able to sufficiently learn the data distribution. However, SN-GAN was able to generate images of flowers that were visually quite similar to the training images, even with only 1,606 training samples. This inspired confidence that conditioning on image categories, despite effectively reducing the amount of training data, could still produce believable images. Results and comparison for flowers can be found in Fig.  \ref{fig:comparison of flowers}. 

SN-GANs additionally had very good performance on landscapes, portraits, and faces. Model performances evaluated using SWD are shown in Table \ref{table:1}, and samples of generated landscapes are shown in Fig. \ref{fig: comparison of landscapes}.

\subsection{Gradient Penalty}

We again performed experiments on all the four image categories, now adding the gradient penalty to the SN loss function. This SNGP model was very expensive to train. As explained in \ref{ssec:method_GP}, the computation of the gradient norm lead to a significant increase in training time. (In our experiments, training time for SNGP-GANs was more than 5 times higher than SN-GANs.) Results better than both DC-GAN and SN-GAN were expected; however, this was not the case. The model outperformed our DC-GAN baseline, but its images were clearly worse than without the gradient penalty for all categories except flowers. We suspect that with more hyperparameter tuning, the model performance could be improved to be at least as good as SN-GAN. However, due to the limited time and computational resources, and given training time needed for SNGP-models, we decided to take SN-GAN as our best model on which to perform conditioning. Comparison of samples generated from the three models can be found in Fig. \ref{fig: comparison of landscapes} and Fig. \ref{fig:comparison of flowers}, while performances evaluated through SWD are in Table \ref{table:1}.

\begin{table}[h!]
\centering
\begin{tabular}{||c c c c||} 
 \hline
Dataset & DC-GAN    & SN &  SNGP      \\ [0.5ex] 
 \hline\hline
landscapes & 69.873  & 37.888 &    42.528  \\
portraits & 117.377 & 58.190&72.032       \\
flowers & 84.299 & 42.457 & 33.973       \\
faces & 67.308 & 35.779 & 42.724       \\ [1ex] 
 \hline
\end{tabular}
\caption{SWD distance x$10^3$. }
\label{table:1}
\end{table}

\subsection{Conditioning}
%Adeel
\subsubsection{Initial Conditioning Experiments}
\label{ssec:experiments-conditioning}
We thought it would be appropriate to start out simple and first condition GANs on MNIST as a proof-of-concept, by replicating the work of \citet{DBLP:journals/corr/MirzaO14}. In this case our $\mathbf{y}$ vector (the conditioning label) was a one-hot encoding of the digit labels with $\mathbf{y}\in\mathbb{R}^{10}$. We trained a simple conditional GAN with no convolutional layers and only a single hidden layer in both the discriminator and generator, as described by \citet{DBLP:journals/corr/MirzaO14}, and the same hyperparameter settings. More details about how the $\mathbf{y}$ vector was introduced to this GAN are in Sec.~\ref{ssec:methodology-conditioning}. After training, we were able to successfully have the generator produce a specific digit by providing it a one-hot encoded $\mathbf{y}$ for the desired digit. Both the generator and discriminator losses were observed to be steady around $1.0$ through the course of training.

We then tried the landscapes + portraits data-subset on the same vanilla conditional GAN architecture, but received very poor results. Both the generator and discriminator losses fluctuated wildly through training, and the generated samples were mostly noise. We attributed this to the fact that the vanilla conditional GAN architecture is too simple for complex images -- it has no convolutional layers which have been repeatedly shown to perform well in image processing in deep learning.

\begin{figure}[h]
\vskip 5mm
\begin{center}
\centerline{\includegraphics[width=0.5\textwidth]{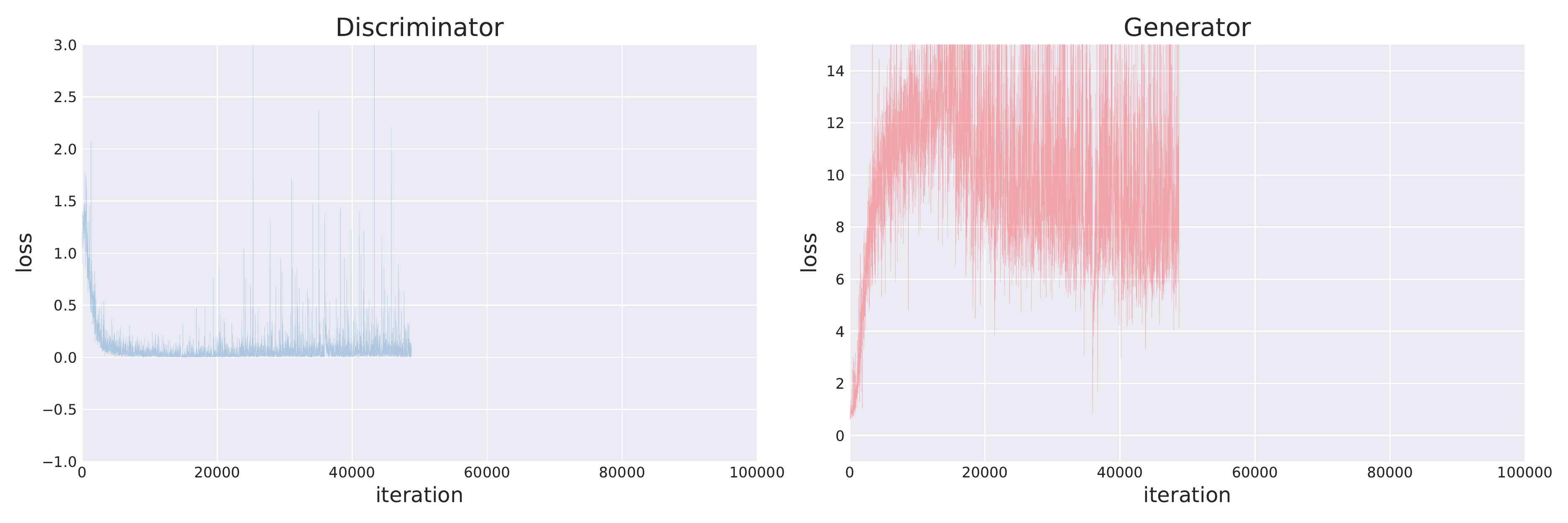}}
\caption{Discriminator and Generator losses plotted over training iterations for a failed experiment with a conditional GAN architecture we tried in earlier experiments. Training was stopped early due to limited time, limited resources, and poor progression of training.}
\label{fig:d_g_losses_failed_faces_conditioning}
\end{center}
\vskip -5mm
\end{figure}

We believed that to get a conditional GAN trained on images more complex than MNIST, we needed convolutional layers in both the discriminator and generator. So we turned to our SN-GAN architecture, which contains convolutional layers in the discriminator and generator (see Fig.~\ref{fig:SNarch}), to add conditioning. Our first conditioning attempt was to start simple and concatenate $\mathbf{y}$ to $\mathbf{z}$ for the input to the generator as in the vanilla conditional GAN, and $\mathbf{y}$ to the dense layer at the end of all the convolutional layers in the discriminator \citep{gauthier2014conditional,reed2016generative}. This did not work. We observed that the discriminator loss would drop to almost $0.0$, while the generator loss would continually increase (to $10.0$ or more), starting from early on in training. Fig.~\ref{fig:d_g_losses_failed_faces_conditioning} shows plots of these D and G losses. The samples generated were those of landscapes and portraits, but of lower quality, and they were randomly placed in our sample grids (as described in Sec.~\ref{ssec:methodology-conditioning}, the first half of our 8x8 samples grid were set to be landscapes and second half to be portraits, which is what we should have observed).

As an alternate approach, we then projected the $\mathbf{y}$ vector through a dense layer, and reshaped it into a fourth channel added to each image being input to the discriminator (Sec.~\ref{ssec:methodology-conditioning}). Furthermore, we decided to keep $\mathbf{y}$ at the end of the discriminator after the convolutional layers as well, in order to ensure the labels reach the end of the discriminator unaltered, and also increase the gradient signal. Hence $\mathbf{y}$ was added both at the beginning and at the end of the discriminator, before and after the convolutional layers. This approach also failed to work for our landscapes + portraits conditioning task. Again, the discriminator loss would drop very close to $0.0$ and generator loss would continually increase (to $10.0$ or more) from early on in the training.

\begin{figure}[h]
\vskip 5mm
\begin{center}
\centerline{\includegraphics[width=0.5\textwidth]{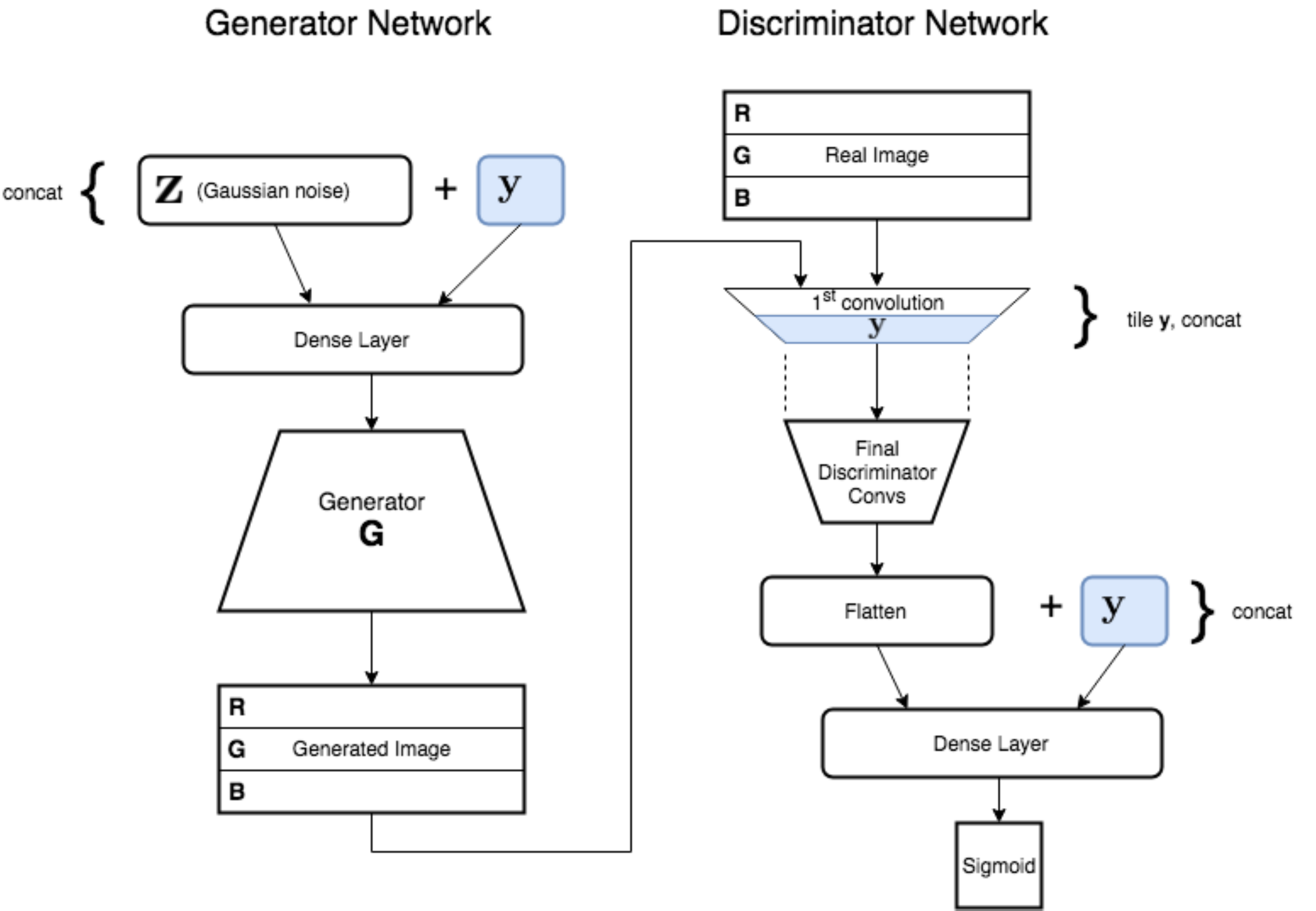}}
\caption{Our conditional GAN architecture.}
\label{fig:mlp_cw4_diagram}
\end{center}
\vskip -5mm
\end{figure}

We then investigated an approach that replicates and then concatenates $\mathbf{y}$ to each filter after the 1st convolutional layer in the discriminator \citep{DBLP:journals/corr/PerarnauWRA16}. We added this to the discriminator in our SN-GAN architecture, again keeping $\mathbf{y}$ at the end of the discriminator after the convolutional layers as well. The generator remained the same, with $\mathbf{y}$ concatenated to the $\mathbf{z}$ noise vector being fed in at the beginning of the generator. A diagram of this conditional GAN architecture can be seen in Fig.~\ref{fig:mlp_cw4_diagram}.

\begin{figure}[h]
\vskip 5mm
\begin{center}
\centerline{\includegraphics[width=0.4\textwidth]{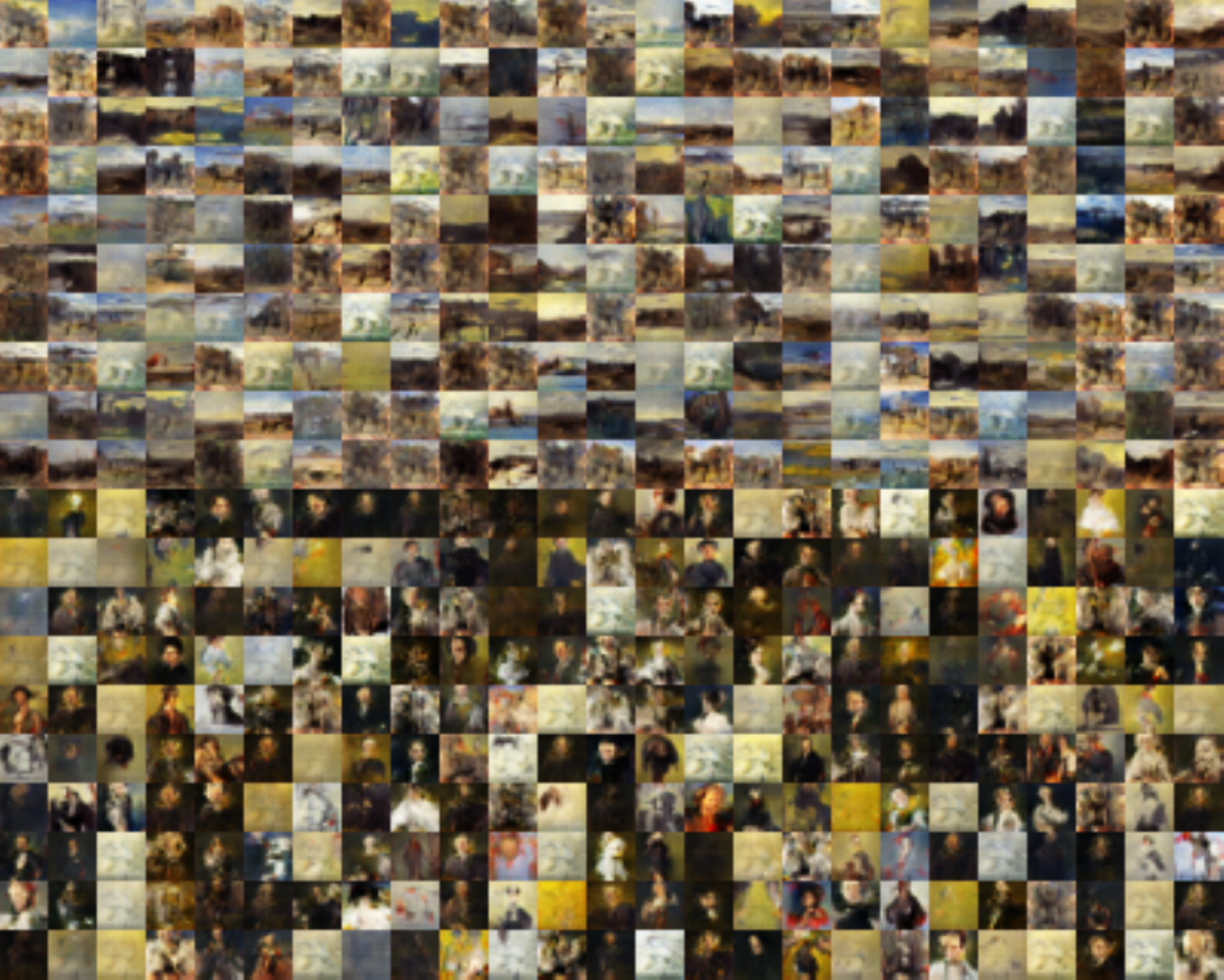}}
\caption{A grid of samples generated using our conditional GAN architecture. $\mathbf{y}$ set to landscapes for top half of the grid, and portraits for the bottom half.}
\label{fig:landcapes-portraits-conditioned}
\end{center}
\vskip -5mm
\end{figure}

\begin{figure}[h]
\vskip 5mm
\begin{center}
\centerline{\includegraphics[width=0.5\textwidth]{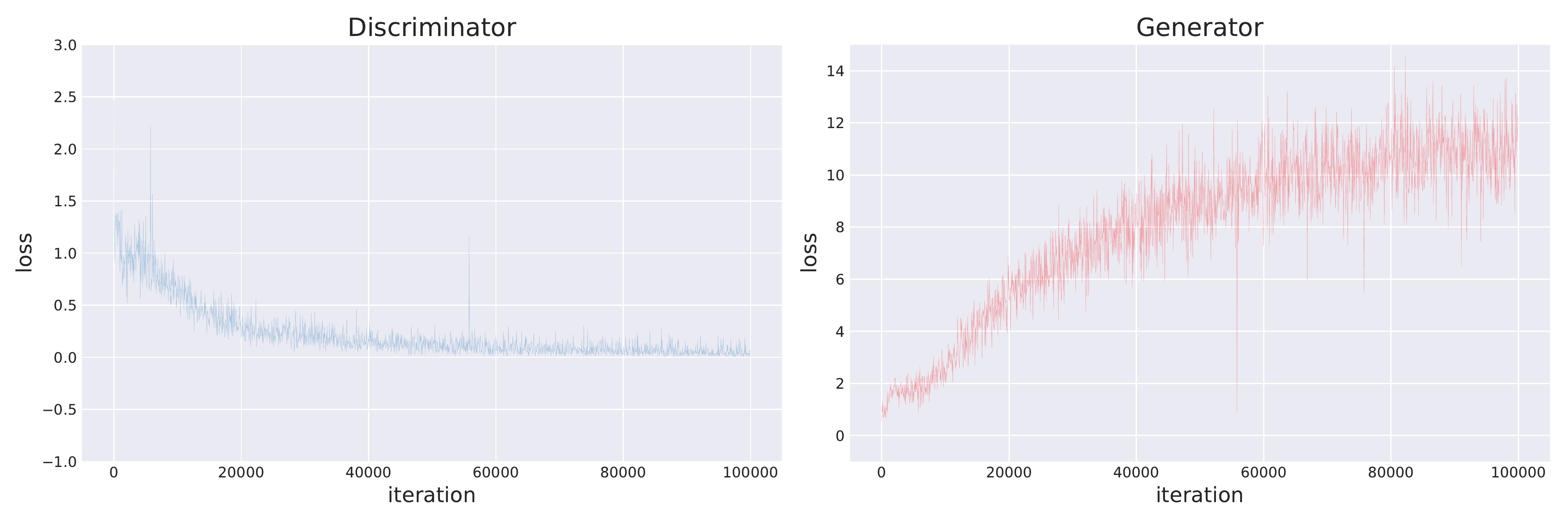}}
\caption{Discriminator and Generator losses plotted over training iterations for a successful experiment with our conditional GAN architecture.}
\label{fig:d_g_losses_successful_faces_conditioning}
\end{center}
\vskip -5mm
\end{figure}

This conditioning architecture worked on the landscapes + portraits conditioning task. The generator was able to produce images conditionally based on $\mathbf{y}$. In this case, our samples grid mostly contained landscapes in the first half, and mostly contained portraits in the second half. An example of this grid (with increased number of samples) can be seen in Fig.~\ref{fig:landcapes-portraits-conditioned}. We also observed that our loss numbers were mostly stable -- they did \emph{slowly} approach $0.0$ for the discriminator and $10.0+$ for the generator, but at a much slower rate and later in training compared to the prior failed conditioning experiments for landscape + portraits. These losses are plotted in Fig.~\ref{fig:d_g_losses_successful_faces_conditioning}. 

\subsubsection{Conditioning on Faces}
\begin{figure}[h]
\vskip 5mm
\begin{center}
\centerline{\includegraphics[width=0.25\textwidth]{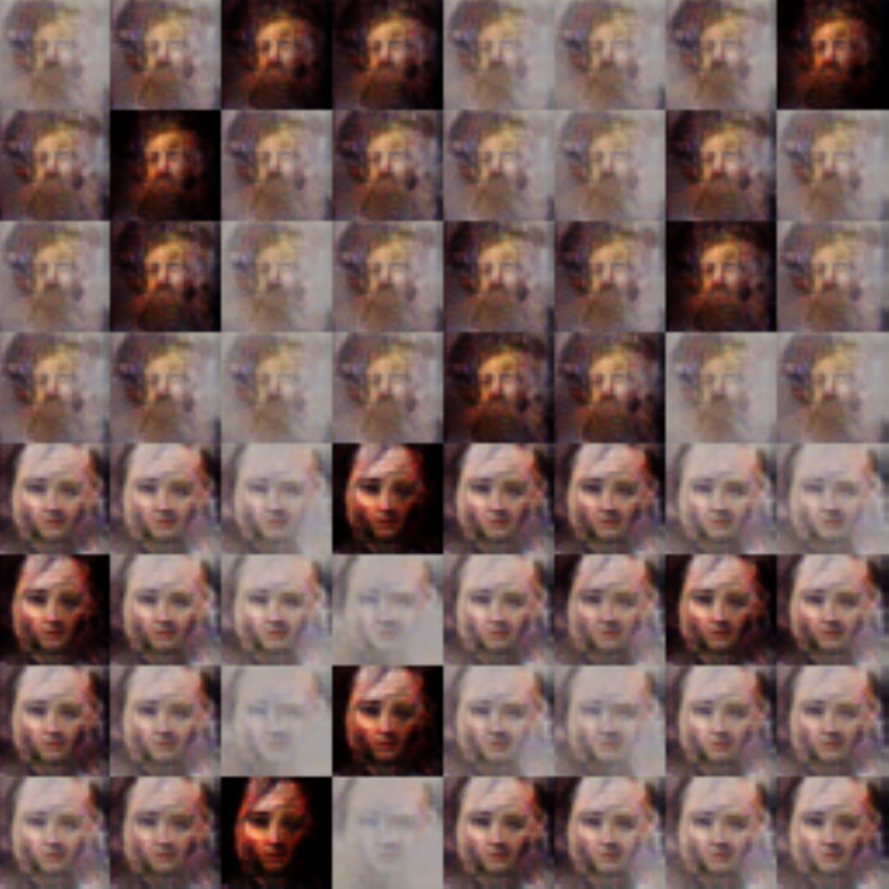}}
\caption{Collapsed conditional GAN being trained on paintings of face images.}
\label{fig:collapsed-conditional-gan-faces}
\end{center}
\vskip -5mm
\end{figure}

Our goal after getting conditioning to work on landscapes + portraits was to try it on the images of faces extracted from all of the portrait paintings in Painter By Numbers. Our $\mathbf{y}$ for each image was a one-hot encoding of facial attributes collected from Face API, as described in Sec.-\ref{ssec:methodology-conditioning}. We started experimenting with conditioning on all 33 facial attributes available to us, but this did not work -- the generated samples throughout training were mostly noise, and loss numbers were highly abnormal compared to those of our successful landscapes + portraits conditioning experiment (Fig.~\ref{fig:d_g_losses_successful_faces_conditioning}). We concluded that this could be due to the fact that there were too many attributes in $\mathbf{y}$, which split our data into too few datapoints available per attribute. In comparison, our successful landscapes + portraits conditioning experiment only had two attributes, and around the same number of data points for each. We narrowed down to 6 attributes for reasons mentioned in our Sec.~\ref{sec:method} -- namely: gender, happiness, age 0-9, black hair, blond hair, facial hair. Training on these, with $\mathbf{y}\in\mathbb{R}^{6}$ instead of $\mathbf{y}\in\mathbb{R}^{33}$, we observed improved results. The generated samples had better quality visually, and the loss numbers were better than the experiment with all 33 attributes. However, the samples did not look as good visually at any point during training as they did when we experimented on generating faces without conditioning. Additionally, the GAN would eventually collapse into a single point before results improved -- see Fig.~\ref{fig:collapsed-conditional-gan-faces}, although it can be observed that the GAN did learn conditioning, as the first half of the samples grid is mostly male and the second half mostly female.

We concluded that our conditioning experiments were failing on the extracted faces for a couple reasons. Paintings of people's faces did not present as much diversity in the distribution of data when compared to landscapes vs portraits. This made for a more difficult task for the conditional GAN. Additionally, we did not have as many datapoints for faces available. We had 15,005 images of landscape paintings, and 16,846 images of portrait paintings, giving us a combined total 31,851 for the landscapes + portraits conditional GAN task, split among 2 attributes. This was in stark contrast to a total of 3,269 images of extracted faces, split among 6 attributes. Hence we decided to augment our faces data-subset, as described in detail in Section~\ref{ssec:methodology-conditioning}. We then set out with 15,190 total augmented images of oil paintings of faces, with conditioning labels (6 attributes per image). We used our same successful conditional GAN architecture from the landscapes + portraits experiment (Fig.~\ref{fig:mlp_cw4_diagram}), and were able to successfully condition on the attributes, but with some caveats. 

For full-sized augmented face images of 128px x 128px, or half-sized augmented face images of 64px x 64px, our conditioning experiments would fail during training. The discriminator loss would quickly get very close to $0.0$, whereas the generator loss would quickly jump up to $10.0+$, and the generator would collapse into a single point, as observed in prior experiments. But when we reduced the image sizes further to 32px x 32px, our experiment succeeded. We noted loss numbers that were more stable than for larger images, and generated samples that were of decent visual quality.

We concluded from these experiments, observing the loss for the discriminator always ending up close to $0.0$, that the discriminator was too powerful and would overfit, making it difficult for the generator to learn anything new. But for smaller images (32px x 32px), there was less to learn in comparison to the larger ones (64px x 64px or greater), so the generator was able to learn the distribution of images quicker, before the discriminator would overfit. So we set out to tune the discriminator to make it less confident, allowing the generator more room to learn to produce better results for conditional generation of larger images.

\citet{salimans2016improved} suggest label smoothing in the discriminator, wherein the positive labels ($D(x) =1$) are replaced with a constant $\alpha < 1$. In our case, we set $\alpha=0.9$. Additionally, \citet{arjovsky2017towards} suggest adding Gaussian noise to the inputs to the discriminator (to both the real images from the training dataset, and fake images from the generator), as covered in Sec. 2. We added zero-mean Gaussian noise with variance $\sigma^2=0.5$. Finally, \citet{salimans2016improved} performed successful GAN training with dropout added to each layer of the discriminator. We added dropout (with rate $0.5$) to each layer, including the input of the discriminator. These improvements resulted in successful training on larger images.
%In future work, some hyperparameter tuning would likely remedy the issue of conditional GAN training on larger images from the augmented faces dataset, perhaps along with techniques such as one-sided label smoothing to make the discriminator less confident \cite{salimans2016improved}.
\begin{figure}[!h]
\centering
\vskip 5mm
\begin{center}
\subfigure [Black haired\newline females] {
	\includegraphics[width=0.14\textwidth]{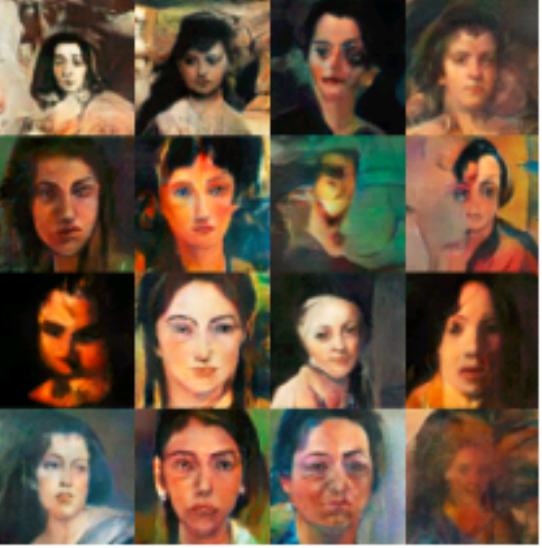}
}
\subfigure[Blond haired\newline females]{
	\includegraphics[width=0.14\textwidth]{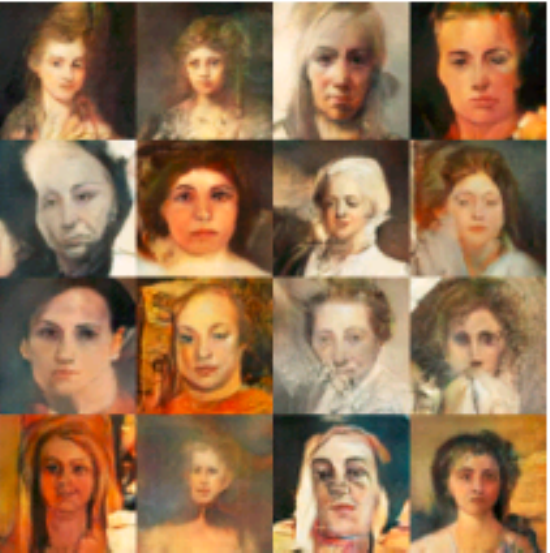}
}
\subfigure[Black haired\newline males with facial hair]{
	\includegraphics[width=0.14\textwidth]{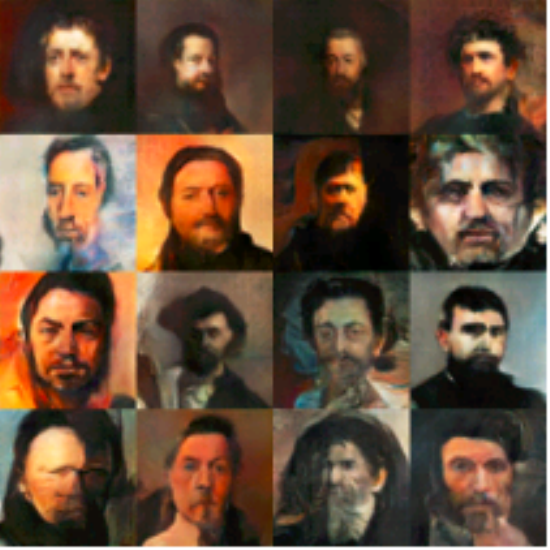}
}
\caption{Samples of faces generated using our conditional GAN architecture trained on images paintings of faces, and facial attribute labels from Microsoft Face API.}
\label{fig:conditioned-faces-samples}   
\end{center}
\vskip -5mm
\end{figure}

Our conditional GAN architecture, with the added adjustments to the discriminator, trained on the augmented faces dataset resized to 64px x 64px, was able to produce visually distinguishable paintings of faces with conditional attributes that were manually specified. When generating a sample, $\mathbf{y}$ could be set to the desired conditional attributes, and passed in with $\mathbf{z}$ to the generator. For example, to produce paintings of blond females, we set $\mathbf{y}=$ [gender=0, happiness=randint(0,1), age\_0-9=randint(0,1), black\_hair=0, blond\_hair=1, facial\_hair=0]. Sample conditional images can be seen in Fig.~\ref{fig:conditioned-faces-samples}. A live demonstration of this pre-trained conditional GAN model can be seen at \url{http://adeel.io/sncgan}.

%Julius
\section{Conclusions}
\label{sec:conclusions}

We performed a series of experiments with the end goal of creating a GAN conditioned on certain image attributes. To do so, we first experimented with several different GAN architectures in order to both learn about the intricacies of their training, and find the optimal architecture on which to apply conditioning. Using paintings of landscapes, flowers, portraits, and extracted faces for training, we compared 3 different modern GAN architectures: Deep Convolutional (DC) GAN, Spectral Normalization (SN) GAN, and Spectral Normalization GAN with Gradient Penalty (SNGP). Both visually and via the Sliced Wasserstein Distance, we found that SN-GAN was the best performing of our models. 

From there we attempted several different approaches found in literature for conditioning GANs on image attributes in a supervised manner. Ultimately, we found a suitable method for attaching the attribute labels (as shown in Fig. \ref{fig:mlp_cw4_diagram}), and created a GAN capable of producing paintings of faces with specified attributes such as gender or hair color. 

This process showed us firsthand the difficulties of training GANs, even with the many recent advances specifically aimed to improve their training. We also saw the incredible power behind deep generative models, despite these difficulties. Research on GANs is exploding as of this writing, and the number of novel tasks to which they can be successfully (and usefully) applied keeps growing. We successfully implemented conditioning on a very modern GAN architecture (and will continue to improve our model and update the website linked above), but more importantly we gained an intimate understanding of some of the intricacies of Generative Adversarial Networks.

\subsection{Future Work}
In future work, it may be useful to perform more extensive hyperparameter tuning to further improve performance. Hyperparameters such as learning rate, $\beta_1$, and $\beta_2$ for Adam, and mini-batch size could be tuned. And Gaussian noise variance, dropout rate, and $\alpha$ for label smoothing could all be further tuned in the discriminator.

Our successful conditioning experiments were performed on 64px x 64px images of paintings of faces. Conditioning on full size images from the same dataset (128px x 128px) could be explored, and may produce good results given the additional adjustments made to the discriminator (Sec.~\ref{ssec:experiments-conditioning}). The same GAN architecture could be tried on real photographs such as those from the CelebA dataset \cite{yang2015facial}, which we predict may yield better results than those for paintings given the consistency of samples in the dataset.

\bibliography{cw4-refs}

\end{document}